\documentclass[final]{cvpr}

\usepackage{times}
\usepackage{epsfig}
\usepackage{graphicx}
\usepackage{amsmath}
\usepackage{amssymb}
\usepackage{multirow}
\usepackage{booktabs}
\usepackage{array}
\usepackage{pdflscape}

\usepackage[pagebackref=true,breaklinks=true,colorlinks,bookmarks=false]{hyperref}

\def\cvprPaperID{1120} 
\def\confYear{CVPR 2021}

\newcommand{\new}[1]{{\color[rgb]{0,0,0}{#1}}}

\newcommand{\para}[1]{\vspace{.05in}\noindent\textbf{#1}}
\def\ie{\emph{i.e.}}
\def\eg{\emph{e.g.}}
\def\etal{{\em et al.}}

\usepackage{cuted}
\usepackage{capt-of}
\begin{document}

\title{Point Cloud Upsampling via Disentangled Refinement}

\author{Ruihui Li \quad Xianzhi Li\thanks{corresponding author} \quad Pheng-Ann Heng \quad Chi-Wing Fu
    \vspace{2mm} \\
	The Chinese University of Hong Kong \\
	{\tt\small \{lirh,xzli,pheng,cwfu\}@cse.cuhk.edu.hk} \vspace{-1cm}
}


\maketitle
\pagestyle{empty}
\thispagestyle{empty}


\begin{strip}\centering
\includegraphics[width=0.9\textwidth]{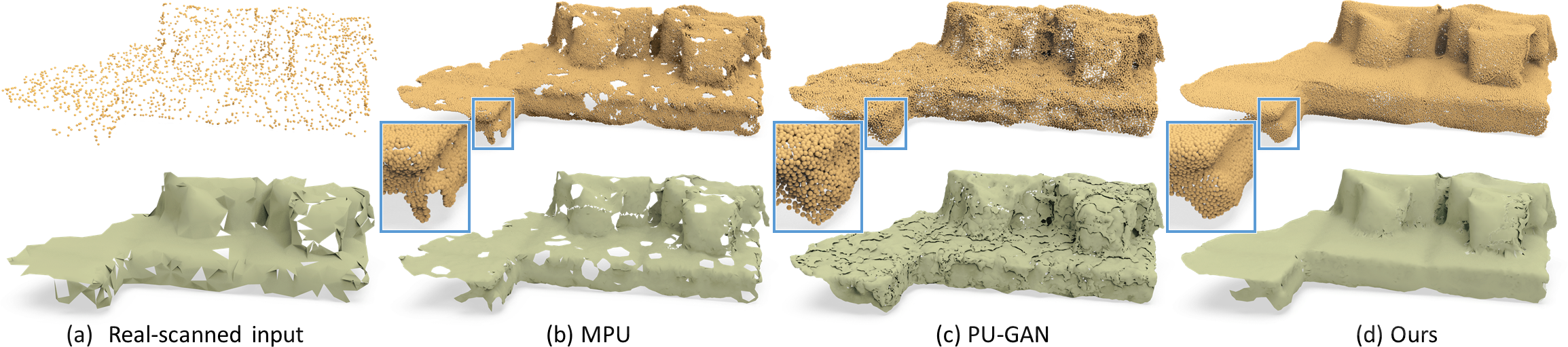}
\captionof{figure}{
In the top row, we show (a) a sparse real-scanned point set from \cite{uy2019revisiting}, followed by upsampled results (16$\times$) produced by (b) MPU~\cite{yifan2018patch}, (c) PU-GAN~\cite{li2019pu}, and (d) our method.
In the bottom row, we show the associated reconstructed 3D meshes produced by the ball-pivoting algorithm~\cite{bernardini1999ball}. Clearly, our method outperforms others on the local uniformity, contributing a better surface reconstruction.
\label{fig:teaser}}
\end{strip}

\begin{abstract}
Point clouds produced by 3D scanning are often sparse, non-uniform, and noisy.
Recent upsampling approaches aim to generate a dense point set, while achieving both distribution uniformity and proximity-to-surface, and possibly amending small holes, all in a single network.
After revisiting the task, we propose to disentangle the task based on its multi-objective nature and formulate two cascaded sub-networks, a dense generator and a spatial refiner.
%
The dense generator infers a coarse but dense output that roughly describes the underlying surface, while the spatial refiner further fine-tunes the coarse output by adjusting the location of each point.
Specifically, we design a pair of local and global refinement units in the spatial refiner to evolve a coarse feature map.
Also, in the spatial refiner, we regress a per-point offset vector to further adjust the coarse outputs in fine scale.
%
Extensive qualitative and quantitative results on both synthetic and real-scanned datasets demonstrate the superiority of our method over the state-of-the-arts.
\new{The code is publicly available at \url{https://github.com/liruihui/Dis-PU}.}
\end{abstract}

\vspace{-5mm}
\section{Introduction}
\label{sec:intro}

Point clouds, as a compact representation of 3D data, are widely explored by both traditional and deep-learning-based methods for many applications~\cite{cole2006using,chen2017multi,orts2016holoportation},~\eg, self-driving cars, robotics, rendering, and medical analysis, etc.
However, raw point clouds produced by 3D scanning are often locally sparse and non-uniform, possibly with small holes on the object surface; see a real-scanned example shown on the top of Figure~\ref{fig:teaser}(a).
%
Obviously, we need to amend such raw data, before we can effectively use it for rendering, analysis, or general processing.


The goal of point cloud upsampling is not limited to generating a dense point set from the sparse input.
Very importantly, the generated points should also faithfully locate on the underlying surface and cover the surface with a uniform distribution.
As an inference-based problem, these goals are very demanding, due to the limited information available in the sparse input.
Besides being sparse, the input points can be non-uniform and noisy, and they may not well represent fine structures (if any) on the underlying surface.

Benefited from data-driven machine learning and deep neural network models, several deep-learning-based methods~\cite{yu2018pu,yu2018ec,yifan2018patch,li2019pu,qian2020pugeo} have been proposed for point cloud upsampling and they demonstrated superior performance, compared with traditional methods~\cite{alexa2003computing,lipman2007parameterization,huang2013edge}.
%
%
The general approach taken in existing learning-based methods is that they first design an upsampling module to expand the number of points in the feature space, then formulate losses to constrain the output points for the distribution uniformity and proximity-to-surface.
In other words, the designed upsampling module is expected not only to infer and generate dense points from the sparse input, but also to produce points that are uniform, clean, and faithfully located on the underlying surface.
However, it is very hard for a network to meet all the requirements at the same time.
Therefore, the dense points produced by existing works still tend to be non-uniform or retain excessive noise (see the top results in Figures~\ref{fig:teaser} (b) \& (c)), thus resulting in low-quality reconstructed meshes (see results in the bottom row).


After revisiting the point cloud upsampling task, we propose to disentangle the task into two sub-goals:
(i) to first generate a {\em coarse but dense point set\/} with less details to roughly describe the underlying surface, and then
(ii) to {\em refine the coarse points\/} to better cover the underlying surface for distribution uniformity and proximity-to-surface.
To do so, we formulate an end-to-end disentangled refinement framework, which consists of two cascaded sub-networks, \emph{a dense generator} and \emph{a spatial refiner}, which are designed to aim for sub-goals (i) and (ii), respectively.
%
%
Particularly, we design the spatial refiner with a pair of local and global refinement units to evolve the coarse feature map inside the refiner to take into account both the local and global geometric structures.
Further, inspired by the residual-learning strategy~\cite{he2016deep}, we formulate the spatial refiner to regress a per-point offset vector for fine-tuning the coarse outputs by adjusting the location of each point.
Compared with directly predicting the final refined 3D point coordinates, regressing a small residual is much easier for the network.

Compared with current upsampling methods~\cite{yu2018pu,yifan2018patch,li2019pu}, our disentangled refinement pipeline assigns lower requirements to each sub-network, so that both the dense generator and the spatial refiner could be more focused on their own sub-goals.
In addition, the cascading scheme allows the two sub-networks to complement each other during network learning, thus leading to a substantial performance boost; see Figure~\ref{fig:teaser}(d).
Extensive experimental results demonstrate that our method
outperforms others on both real-scanned and synthetic inputs.

\section{Related Work}
\label{sec:bg}

\para{Optimization-based upsampling.} \
To generate new points from the inputs, optimization-based methods typically rely on hand-crafted priors.
Alexa~\etal~\cite{alexa2003computing} introduced an early work that inserts new points at the vertices of the Voronoi diagram, which is computed based on the moving-least-squares surface.
Later, a locally optimal projection operator was proposed by Lipman~\etal~\cite{lipman2007parameterization}, where points are resampled based on the $L_1$ norm.
To upsample point cloud in an edge-aware manner, Huang~\etal~\cite{huang2013edge} proposed to first upsample points away from the edges then progressively move points towards the edge singularities.
Later, Wu~\etal~\cite{wu2015deep} introduced a point-set consolidation method by augmenting surface points into deep points that lie on the meso-skeleton of the shape.
Overall, optimization-based methods may fail when the prior assumptions are not satisfied.

\para{Deep learning-based upsampling.} \
Inspired by the success of PointNet~\cite{qi2017pointnet}, many deep learning methods were proposed for assorted tasks on point cloud processing, from high-level tasks like classification~\cite{li2018pointcnn,zhang2019shellnet,li2020rotation} and object detection~\cite{lang2019pointpillars,qi2019deep} to low-level tasks like completion~\cite{yuan2018pcn,chen2020unpaired},
denoising~\cite{rakotosaona2019pointcleannet,hermosilla2019total}, and other applications~\cite{wang2019deep,chen2020lassonet,li2020pointaugment,li2020unsupervised}.

For the point cloud upsampling task, Yu~\etal~\cite{yu2018pu} introduce PU-Net, the first attempt based on deep learning to learn to extract multi-scale features and expand a point set via a multi-branch convolution in the feature space.
Later, they introduce EC-Net~\cite{yu2018ec} to achieve edge-aware point cloud upsampling, which further enhances the surface reconstruction quality.
Soon after that, Wang~\etal~\cite{yifan2018patch} proposed MPU, a network that progressively upsamples point patches in multiple steps, while
Li~\etal~\cite{li2019pu} proposed PU-GAN by leveraging the generative adversarial network to learn to synthesize points with a uniform distribution in the latent space.
Very recently, Qian~\etal~\cite{qian2020pugeo} proposed PUGeoNet to first generate samples in a 2D domain and then use a linear transform to lift up the samples to 3D.

\begin{figure*}[t]
\centering
\includegraphics[width=0.98\linewidth]{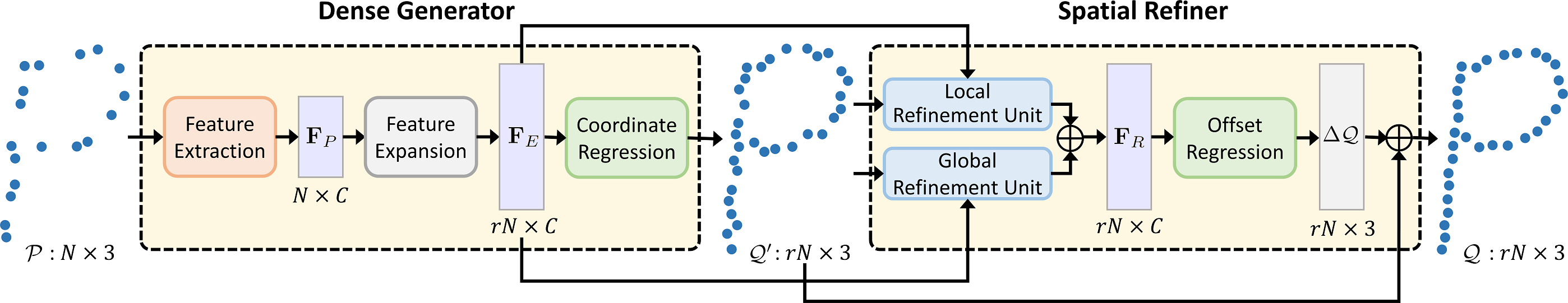}
\caption{An illustration of our framework.
Given sparse input $\mathcal{P}$ of $N$ points, the dense generator extracts feature map $\mathbf{F}_P$ from the input, generates the expanded feature map $\mathbf{F}_E$, then produces a coarse but dense point set $\mathcal{Q}'$ of $rN$ points, where $r$ is the upsampling rate.
Next, the spatial refiner consumes both $\mathcal{Q}'$ and the associated $\mathbf{F}_E$ to obtain the refined feature map $\mathbf{F}_R$ via a pair of local and global refinement units. We then regress offsets $\Delta \mathcal{Q}$ from $\mathbf{F}_R$, and output the final refined dense points $\mathcal{Q}$ by $\mathcal{Q}'+\Delta \mathcal{Q}$.}
\label{fig:framework}
\vspace*{-3mm}
\end{figure*}

After revisiting deep-learning-based methods for point cloud upsampling, we found that existing works all rely on a single network to meet all the various goals of point cloud upsampling,~\ie, dense point set generation, faithfulness to underlying surface, distribution uniformity, hole amendment, etc.
To better meet the multi-objective nature of the task, we propose a new approach to disentangle the task into two cascaded networks and demonstrate substantial improvements over the prior works.


\section{Method}
\label{sec:method}

\subsection{Overview}
\label{subsec:overview}

Essentially, 3D scanning is a sampling problem in the 3D physical space, while upsampling is a prediction problem, aiming to infer more samples on the original surface, given the sparse samples obtained in the scanning.
Given a point set $\mathcal{P}$ of $N$ points, which is typically sparse, non-uniform, and noisy, the point cloud sampling task aims to generate a dense point set, say $\mathcal{Q}$ of $rN$ points, for a given upsampling rate $r$.
These upsampled points should
(i) faithfully describe the underlying object surface and
(ii) cover the surface with a uniform distribution.
This upsampling task is very challenging, since we need to infer new knowledge from the sparse input, in which the information about the original geometry is not completely represented.
%

Different from the existing approaches that try to meet the various goals in upsampling all in a single network, we propose to disentangle the upsampling task into two sub-goals, where we first generate a coarse but dense point set and then refine these points over the underlying surface to improve the distribution uniformity.
Before giving the details of our method, we first discuss our key insights:
\begin{itemize}
\vspace*{-1mm}
\item
First, we propose an end-to-end disentangled refinement framework with two cascaded sub-networks: one to generate dense points that roughly locate on the underlying surface and the other to aim for proximity-to-surface and distribution uniformity.
Thus, each sub-network can better focus on its specific sub-goal, while complementing each other in the upsampling task.
\vspace*{-1mm}
\item
Second, with the help from the spatial refiner sub-network, the dense generator sub-network does not need a complicated structure.
Hence, having a simple yet effective structure can enable it to expand features with higher flexibility and increase the upsampling rate without introducing extra network parameters.
%
\vspace*{-1mm}
\item
Third, we design the spatial refiner with both local and global refinements, such that we can leverage their complementary strengths to locally improve the distribution uniformity and proximity-to-surface, and globally explore similar structures on the surface.
%
%
\end{itemize}

Figure~\ref{fig:framework} shows the overall framework of our method.
Given a sparse point set $\mathcal{P}$, we first feed it into our \emph{dense generator} to generate the dense output $\mathcal{Q}'$ with $rN$ points.
At present, $\mathcal{Q}'$ may still be non-uniform and noisy like $\mathcal{P}$, as illustrated in the figure's toy example.
Next, we feed it into our \emph{spatial refiner} to further regress a per-point offset vector $\Delta \mathcal{Q}$, which is used to adjust the location of each point in $\mathcal{Q}'$, such that the refined dense points $\mathcal{Q}$ can faithfully locate on the underlying surface, while being more uniform.
In the following, we first present the details of the dense generator and spatial refiner in Sections~\ref{subsec:dense} and~\ref{subsec:refiner}, respectively. Then, we give the details of the patch-based end-to-end network training in Section~\ref{subsec:train}.


\subsection{Dense Generator}
\label{subsec:dense}

Given $\mathcal{P} \in \mathbb{R}^{N \times 3}$, our dense generator produces the upsampled coarse points $\mathcal{Q}' \in \mathbb{R}^{rN \times 3}$.
Similar to existing upsampling approaches~\cite{yu2018pu,yifan2018patch,li2019pu}, we also expand the number of points in the feature space.

Specifically, as shown in the left-side of Figure~\ref{fig:framework}, we first employ a feature extraction unit to embed the feature map $\mathbf{F}_P \in \mathbb{R}^{N \times C}$ from $\mathcal{P}$, where $C$ is the number of feature channels.
Here, we follow \cite{yifan2018patch} to use the same feature extraction unit by considering the efficiency and effectiveness.
Please refer to \cite{yifan2018patch} for the details of this unit.
Next, we feed $\mathbf{F}_P$ into a feature expansion unit to generate the expanded feature map $\mathbf{F}_E \in \mathbb{R}^{rN \times C}$.
As discussed in Section~\ref{subsec:overview}, with the help from the cascaded spatial refiner, the dense generator only needs to generate a dense point set to roughly locate on the underlying surface.
Hence, in the feature expansion unit, we adopt the commonly-used expansion operation by duplicating $\mathbf{F}_P$ with $r$ copies and concatenating with a regular 2D grid to obtain $\mathbf{F}_E$.
Although such operation may introduce redundant information or extra noise, these problems could be rectified by the subsequent spatial refiner.
Lastly, $\mathcal{Q}'$ is generated by regressing the point coordinates from $\mathbf{F}_E$  via multi-layer perceptrons (MLPs).

\begin{figure*}[t]
	\centering
	\includegraphics[width=0.85\linewidth]{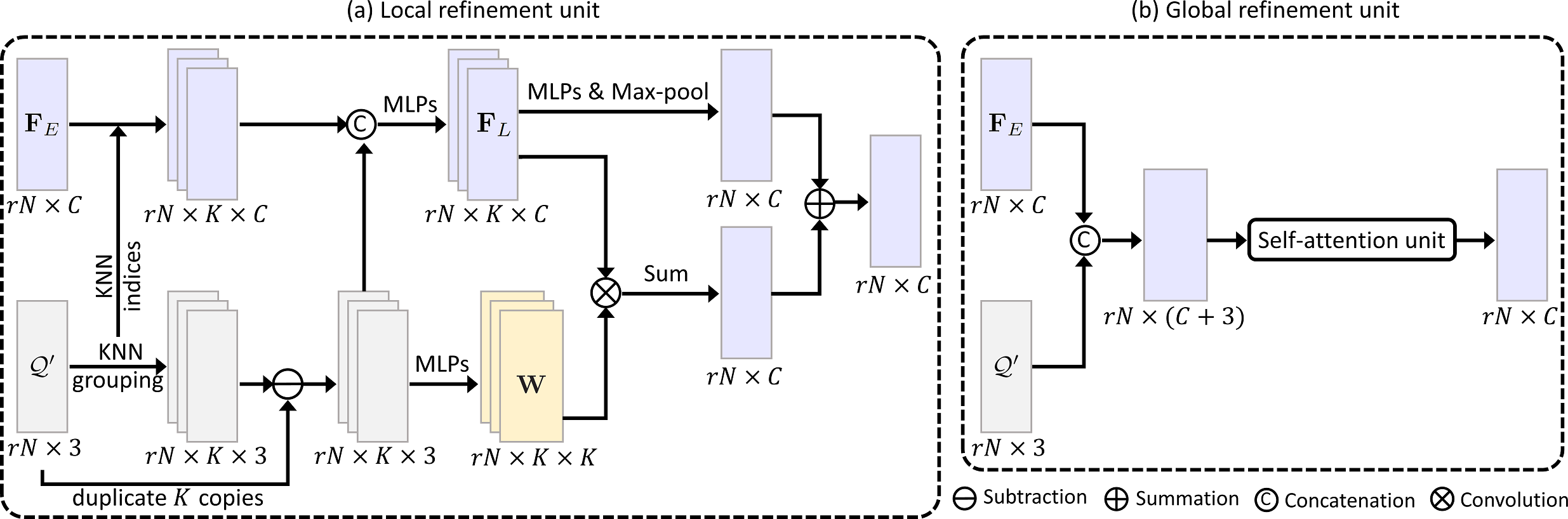}
	\caption{The architecture of (a) the local refinement unit and (b) the global refinement unit.}
	\label{fig:local}
	\vspace*{-3mm}
\end{figure*}

\subsection{Spatial Refiner}
\label{subsec:refiner}
Considering that $\mathcal{Q}'$ may still be noisy and non-uniform, we thus design a spatial refiner to further fine-tune the spatial location of each point in $\mathcal{Q}'$ and generate a high-quality dense point set $\mathcal{Q}$, which lies on the underlying surface and also distributes uniformly.

To do so, as shown in the right-side of Figure~\ref{fig:framework}, we first feed the coarse $\mathcal{Q}'$ and the associated coarse feature map $\mathbf{F}_E$ into both local and global refinement units, which are detailed later.
Next, we sum the two outputs generated by the two refinement units to obtain the refined feature map $\mathbf{F}_R \in \mathbb{R}^{rN \times C}$.
Then, instead of directly regressing the refined point coordinates, we adopt residual learning to regress the per-point offset $\Delta \mathcal{Q}$.
The reason behind is that, compared with the relative offset, the absolute point coordinates are more diverse and have a wide distribution in 3D space.
Hence, it is difficult for the network to synthesize points without introducing extra noise, while still preserving the uniformity and shape structures.
Lastly, the final output $\mathcal{Q}$ is obtained by $\mathcal{Q}'+\Delta \mathcal{Q}$.

\para{Local refinement unit} aims to evolve $\mathbf{F}_E$ by considering the local geometric structures.
Figure~\ref{fig:local}(a) shows the detailed architecture.
Specifically, we first employ KNN grouping on $\mathcal{Q}'$ to search $K$-nearest neighbors and group the associated neighbor points together to obtain a stacked $rN \times K \times 3$ point volume.
At the same time, we employ the same nearest neighbor indices to group $\mathbf{F}_E$ into an $rN \times K \times C$ feature volume.
Then, we duplicate $\mathcal{Q}'$ with $K$ copies and apply a subtraction operation on the duplicated and the grouped point volumes, which helps encode local information.
We then concatenate the subtracted point volume with the grouped feature volume and apply MLPs to obtain the encoded local feature volume $\mathbf{F}_L \in \mathbb{R}^{rN \times K \times C}$.

Next, to obtain the local point feature over $\mathbf{F}_L$, a common routine is to apply MLPs followed by a max-pooling along the $K$-dimension.
However, to account for the relative importance among the $K$ neighbors, we further regress a spatial weight $\mathbf{W}$ (see the light yellow volume in Figure~\ref{fig:local} (a)) from the subtracted point volume.
Then, we modify $\mathbf{F}_L$ via a convolution with $\mathbf{W}$, followed by a summation along the $K$-dimension to obtain the weighted $rN \times C$ feature map.
Lastly, we sum the weighted feature map as the final refined local features.

\para{Global refinement unit} aims to refine $\mathbf{F}_E$ by considering the overall shape structure.
As shown in Figure~\ref{fig:local}(b), instead of feeding only $\mathbf{F}_E$ to the refinement unit, we concatenate $\mathbf{F}_E$ and $\mathcal{Q}'$ together as the input to avoid losing the overall shape structure.
Next, we adopt the widely-used self-attention unit~\cite{zhang2019self} to obtain the refined global feature map, since this unit regresses attention weights among all the $rN$ points, thus introducing long-range context dependencies.
For brevity, we will not describe the details of this attention unit; please refer to \cite{zhang2019self} if needed.

\subsection{Patch-based End-to-end Training}
\label{subsec:train}

Since point cloud upsampling is a low-level task that requires us to focus more on the local geometric structures, we thus adopt the patch-based training strategy, as all the existing upsampling approaches did.
During training, for each input sparse point set $\mathcal{P}$ and its associated target dense point set $\mathcal{\hat{Q}}$, our framework predicts both $\mathcal{Q}'$ and $\mathcal{Q}$.
Hence, we formulate our objective function to encourage the geometric consistency between $\mathcal{Q}'$\&$\mathcal{\hat{Q}}$, and between $\mathcal{Q}$\&$\mathcal{\hat{Q}}$:
\begin{equation}
\label{equ:loss}
	\mathcal{L} = \mathcal{L}_{\text{CD}}(\mathcal{Q}', \mathcal{\hat{Q}}) + \lambda \mathcal{L}_{\text{CD}}(\mathcal{Q}, \mathcal{\hat{Q}}),
\end{equation}
where $\mathcal{L}_{\text{CD}}(\cdot)$ means the Chamfer Distance (CD)~\cite{fan2017point} to measure the average closest point distance between two point sets.
The parameter $\lambda$ controls the relative importance of each term.
In the early stage of network training, we set a small $\lambda$, so that the network focuses more on the training of the dense generator to produce a more reliable $\mathcal{Q}'$.
As the training progresses, we gradually increase $\lambda$ to let our spatial refiner to be fully trained.

Note that, we also tried to combine Eq.~\eqref{equ:loss} with the repulsion loss~\cite{yu2018pu} to encourage the distribution uniformity.
However, we found that this repulsion loss does not contribute too much in our work, because our method can already generate a relatively uniform dense point set benefited from the disentangled refinement scheme, even without any losses to constrain the uniformity distribution.


\section{Experiments}
\label{sec:experiment}

\subsection{Experimental Settings}
\label{subsec:datasets}

\para{Datasets.} \
We employ both synthetic and real-scanned datasets in our experiments.
For the synthetic dataset, we use the benchmark dataset provided by~\cite{li2019pu} with 120 training and 27 testing objects.
For each training object, we follow~\cite{li2019pu} to crop 200 overlapped patches, thus resulting in totally 24,000 training surface patches. 
On each surface patch, we uniformly sample $rN$ points as target $\mathcal{\hat{Q}}$, and then randomly downsample $N$ points from $\mathcal{\hat{Q}}$ as the training input $\mathcal{P}$.
For each testing shape, we follow MPU~\cite{yifan2018patch} and PU-GAN~\cite{li2019pu} to sample $\sim$20,000 uniform points using Poisson disk sampling as $\mathcal{\hat{Q}}$ for quantitative evaluation, and generate 1,024 non-uniform points for testing.

For real-scanned dataset, we use ScanObjectNN~\cite{uy2019revisiting}, which contains 2,902 point cloud objects in 15 categories.
Each object has 2,048 points.
Since no target dense points are provided in ScanObjectNN, we just use this dataset for testing.
Hence, during testing, we have both synthetic and real-scanned point clouds.
For each testing point cloud with 2,048 points, we use farthest sampling to pick seeds and extract a local patch of $N$ points per seed.
We then feed these patches to the network for testing and combine the upsampled results as the final output.

%

\para{Evaluation metrics.} \
For quantitative evaluation, we consider three widely-used evaluation metrics:
(i) Chamfer distance (CD),
(ii) Hausdorff distance (HD), and
(iii) Point-to-surface (P2F) distance using the original testing objects.
A lower evaluation metric indicates a better performance.

\para{Comparison methods.} \
To demonstrate the effectiveness of our method, we compare it with three state-of-the-art point cloud upsampling methods, including PU-Net~\cite{yu2018pu}, MPU~\cite{yifan2018patch}, and PU-GAN~\cite{li2019pu}.
We use their released public code and follow the same setting in the original papers to re-train their networks using our prepared training data.
Note that, for the recent work PUGeoNet~\cite{qian2020pugeo}, we cannot provide the comparison results without available code so far.

\begin{figure*}[htbp]
	\centering
	\includegraphics[width=0.94\linewidth]{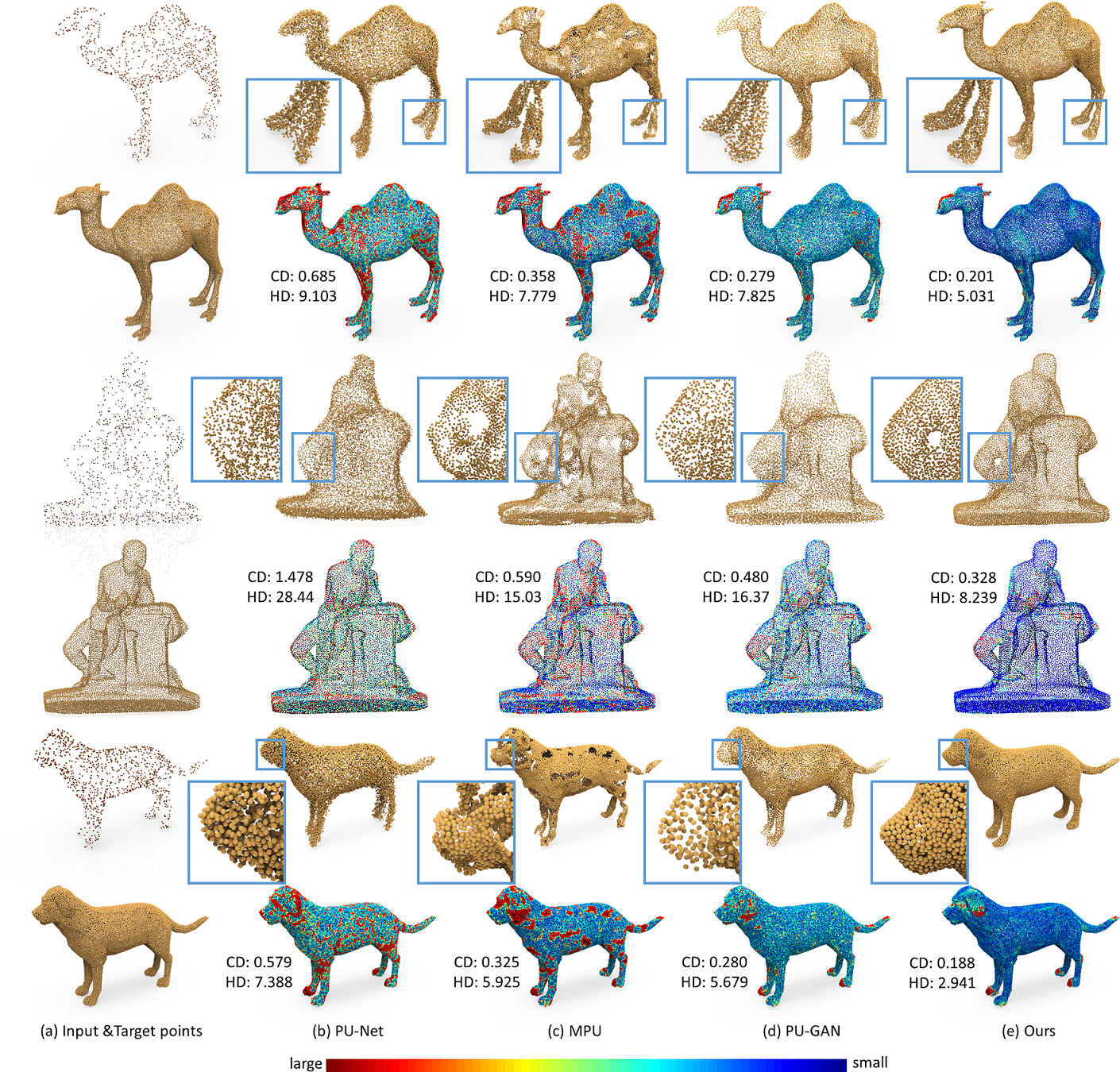}
	\caption{Comparing point set upsampling (x16) results from synthetic sparse inputs (a) using different methods (b-e). We also show the associated error maps, where the colors reveal the nearest distance for each target  point to the predicted point set generated by each method.}
	\label{fig:qualitativeSyn}
	\vspace*{-4mm}
\end{figure*}


\para{Implementation details.} \
In experiments, we set $N=256$.
We train our network with a batch size of 28 for 400 epochs on the TensorFlow platform.
For each patch, we apply random scaling, rotation, and point perturbation to avoid over-fitting.
The Adam optimizer is used with the learning rate of 0.001, which is linearly decreased by a decay rate of 0.7 per 40 epochs until $10^{-6}$.
The parameter $\lambda$ in Eq.~\eqref{equ:loss} is linearly increased from 0.01 to 1.0 as the training progresses.

\subsection{Results on Real-scanned Dataset}
\label{subsec:real}
\begin{figure*}[t]
	\centering
	\includegraphics[width=0.99\linewidth]{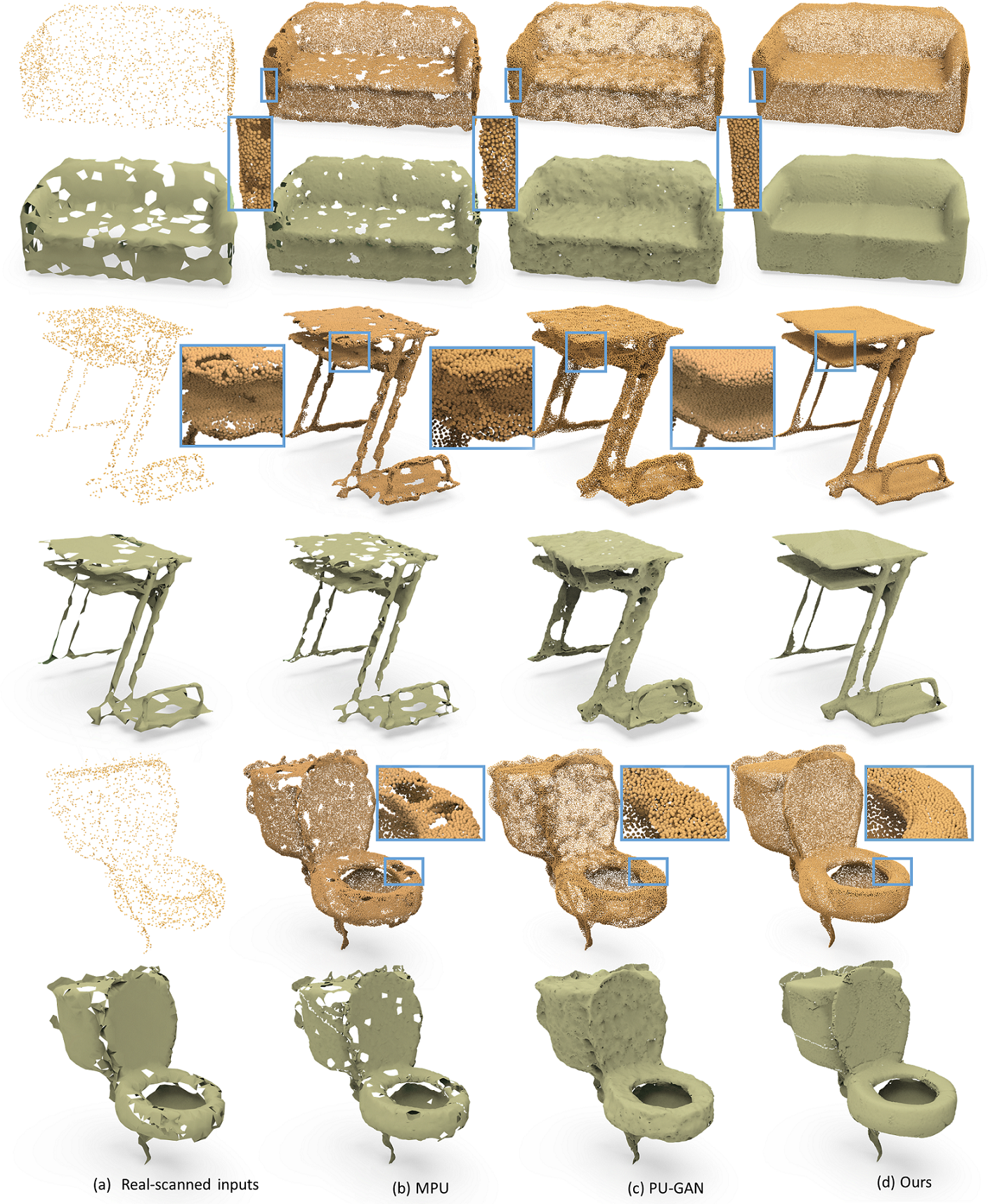}
	\caption{Comparing point set upsampling (16$\times$) results and reconstructed 3D meshes using different methods (b-d) from real-scanned sparse inputs (a), while the bottom row shows the reconstructed meshes. Clearly, our method outperforms others on the local uniformity, contributing a better surface reconstruction.}
	\label{fig:qualitativeScan}
\end{figure*}

First, we compared our method with state-of-the-arts on real-scanned test inputs.
Besides the results shown earlier in Figure~\ref{fig:teaser}, we further show more visual comparison results in Figure~\ref{fig:qualitativeScan} (see page 8), where we set $r=16$.
For each object, the top row shows the upsampled points by each method, and the bottom row shows the associated reconstructed 3D meshes using ball-pivoting surface reconstruction algorithm~\cite{bernardini1999ball}.
Note that, since PU-Net is an early work with not very promising results, we thus omit its results in visual comparisons.
As shown in the top row of Figure~\ref{fig:qualitativeScan}(a), to upsample the real-scanned sparse inputs is very challenging, since these points are not only noisy and non-uniform, but also exhibit many small holes and structural defects.
Thus, reconstructing meshes directly from sparse inputs often results in incomplete surfaces with many holes; see the bottom results in (a).
Comparing the upsampled points produced by various methods, the other methods tend to retain noise in their results, or fail to generate a uniform output, thus resulting in low-quality reconstructed meshes with small holes or rough surfaces.
On the contrary, our method enables to produce uniform dense points with low deviations to the underlying object surface.
Hence, the reconstructed meshes from our upsampled points can well describe the geometric structures with smooth and complete surfaces.
More real-scanned comparisons can be found in our supplementary material.

\begin{table}[t]
	\centering
	\caption{Quantitative comparisons by using our method and state-of-the-arts.
		The units of CD, HD, and P2F are all $10^{-3}$.
	}
	\label{tab:quanComparison}
	\resizebox{\linewidth}{!}{
		\begin{tabular}{@{\hspace{1mm}}c@{\hspace{1mm}}||
				@{\hspace{1mm}}c@{\hspace{1mm}}||c@{\hspace{3mm}}c@{\hspace{3mm}}c@{\hspace{1mm}} ||
				@{\hspace{1mm}}c@{\hspace{1mm}}||c@{\hspace{3mm}}c@{\hspace{3mm}}c@{\hspace{1mm}}
			} \toprule[1pt]
			\multirow{2}*{Methods}
			& \multicolumn{4}{@{\hspace{1mm}}c@{\hspace{1mm}}||@{\hspace{1mm}}}{4X}
			& \multicolumn{4}{@{\hspace{1mm}}c@{\hspace{1mm}}@{\hspace{1mm}}}{16X} \\
			
			\cline{2-5} \cline{6-9}
			& Size & CD & HD & P2F & Size & CD & HD & P2F\\ \hline \hline
			PU-Net~\cite{yu2018pu}
			&10.1M   & 0.844  & 7.061  & 9.431
			&24.5M   & 0.699  & 8.594  & 11.619      \\
			MPU~\cite{yifan2018patch}
			&23.1M & 0.632  & 6.998  & 6.199
			&92.5M & 0.348  & 7.187  & 6.822      \\
			PU-GAN~\cite{li2019pu}
			&9.57M & 0.483  & 5.323  & 5.053
			&9.57M & 0.269  & 7.127  & 6.306    \\ \hline
			Our
			&13.2M & \textbf{0.315} & \textbf{4.201} & \textbf{4.149}
			&13.2M & \textbf{0.199} & \textbf{4.716} & \textbf{4.249}
			\\ \bottomrule[1pt]
	\end{tabular}}
	\vspace*{-3mm}
\end{table}

\subsection{Results on Synthetic Dataset}
\label{subsec:synthetic}
Next, we compared our method with state-of-the-arts on synthetic test models provided by~\cite{li2019pu}.
Figure~\ref{fig:qualitativeSyn} shows the visual comparisons on three sparse inputs, where we set $r$$=$$16$.
Comparing the dense points produced by our method (e) and others (b-d) with the target (a), we can see that other methods tend to introduce excessive noise (\eg, (b)), cluster points together with a non-uniform distribution (\eg, (c)), or destroy some tiny structures (\eg, (d)) in the results.
In contrast, our method produces the most similar visual results to the target points, and our dense points can well preserve tiny local structures with a uniform point distribution; see particularly the blown-up view in Figure~\ref{fig:qualitativeSyn}.
Besides, we show also the associated error maps, where the colors reveal the nearest distance for each point in target point set to the predicted point set.
We can see that the errors of our upsampled results are the lowest (\ie, most points are blue), which is also verified by both CD and HD values.
More comparison results can be found in the supplemental material.

Table~\ref{tab:quanComparison} shows the quantitative comparisons on all the synthetic test models under different upsampling rates.
We can see that our method achieves the lowest values on all the evaluation metrics in terms of both upsampling rates.
Note that, different from PU-Net and MPU, the number of learnable parameters in our network will not increase as $r$ increases.
Hence, our method has good scalability to a large upsampling rate.
More importantly, when $r$ increases, the advantages of our method compared with others become more obvious.
The reason behind is that, the prediction difficulty will significantly increase given a large $r$ for existing approaches.
However, thanks to the disentangled refinement scheme, our method has better adaptability.

\begin{figure*}[t]
	\centering
	\includegraphics[width=0.95\linewidth]{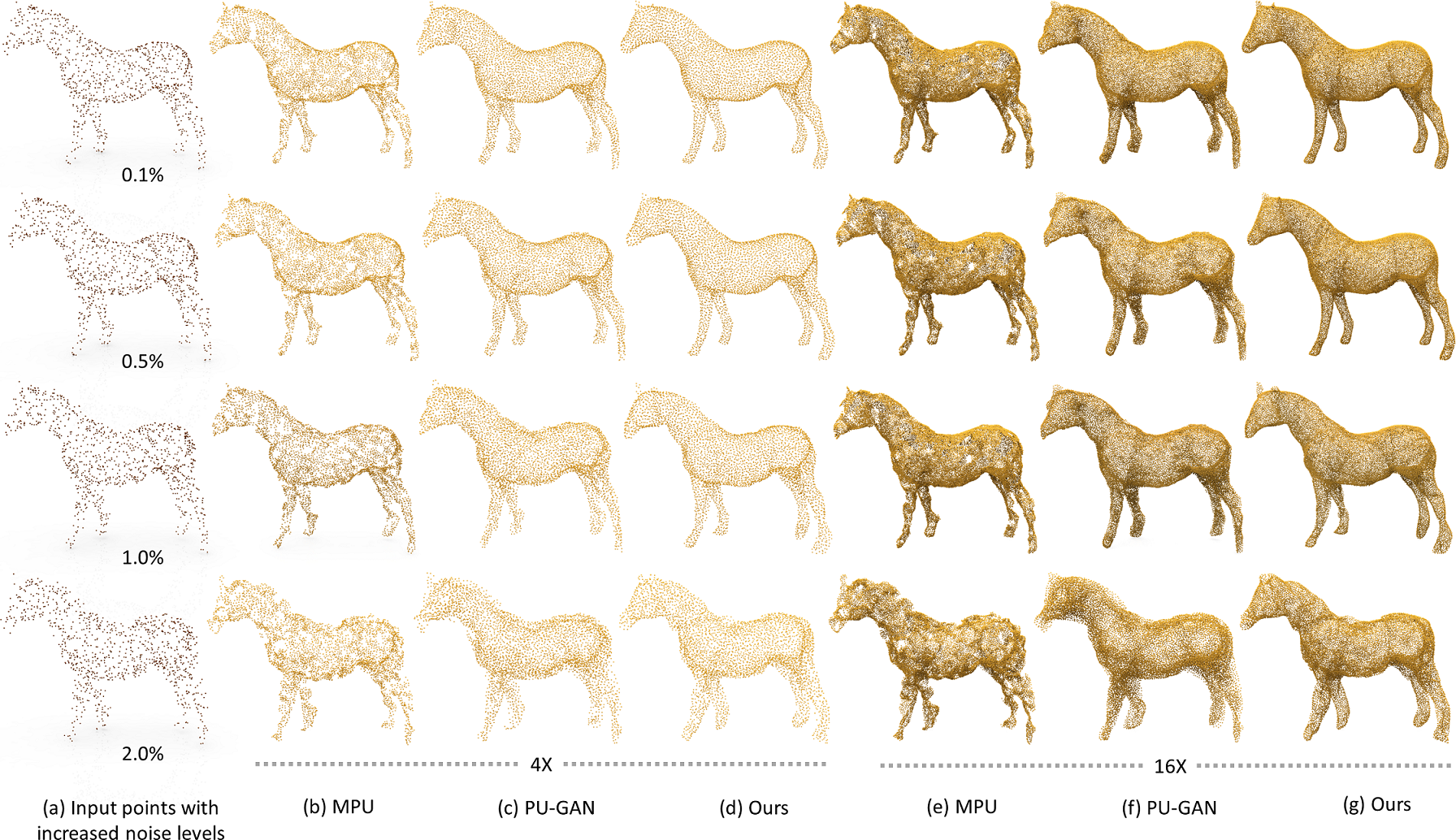}
	\caption{Comparing point set upsampling results produced using different methods under different upsampling rate, when given noisy sparse inputs with increasing noise level,~\ie, 0.1\%, 0.5\%, 1.0\%, and 2.0\%.}
	\label{fig:noise}
	\vspace*{-1mm}
\end{figure*}

\subsection{Noise Robustness Test}
\label{subsec:noise}
We explored the noise robustness of our method by adding Gaussian noise of different levels to the synthetic test inputs.
Figure~\ref{fig:noise} shows the visual comparisons. 
Clearly, our method achieves more uniform upsampling results (d \& g) without excessive noise, under both upsampling rates.
The quantitative comparisons are summarized in Table~\ref{tab:noise}, where $r=16$ and we show the CD values.
Obviously, our method produces the lowest values across all the noise levels with a significant margin, compared to others.

\vspace*{-2mm}
\subsection{Ablation Study}
\label{subsec:ablation}
\begin{figure}
\centering
\includegraphics[width=0.95\linewidth]{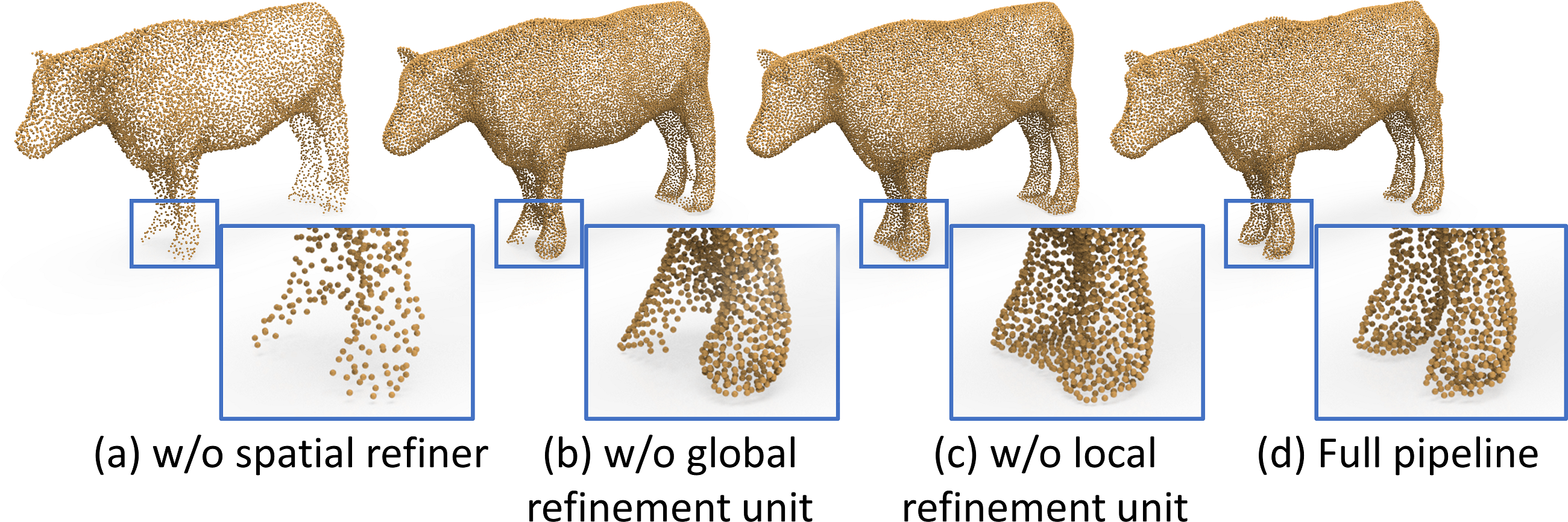}
\caption{Visualization results for the ablation study.}
\label{fig:ablation}
\vspace*{-4mm}
\end{figure}

To evaluate the effectiveness of the major components in our framework, we conducted an ablation study by simplifying our full pipeline in the following four cases: (A) removing the spatial refiner and only keeping the dense generator (see Figure~\ref{fig:framework}); (B) removing the local refinement unit; (C) removing the global refinement unit; and (D) removing offsets and directly regressing the point coordinates.
In each case, we re-trained the network and tested the performance using synthetic data.
Table~\ref{tab:ablation} summarizes the results of each case in terms of CD value, compared to our full pipeline (bottom row).
Clearly, our full pipeline performs the best with the lowest CD value, and removing any component reduces the overall performance, meaning that each component in our framework contributes.
\new{We also present a visual result associated with the ablation study in Figure~\ref{fig:ablation}.
The supplemental material provides more results with various settings, verifying the effectiveness of our disentangled design.
Noted that since the dense generator adopts an identical or simplified design from the previous works~\cite{yifan2018patch,li2019pu}, our spatial refiner is generally applicable to other networks.}
%

\begin{table}
	\centering
	\caption{Quantitative comparisons by using our method and state-of-the-arts to upsample noisy inputs with increasing noise level ($r=16$). Here we show CD values with the unit of  $10^{-3}$.}
	\label{tab:noise}
    \vspace{2mm}
	\resizebox{0.9\linewidth}{!}{
		\begin{tabular}{@{\hspace{1mm}}c@{\hspace{1mm}}||
				@{\hspace{1mm}}c@{\hspace{3mm}}c@{\hspace{3mm}}c@{\hspace{3mm}}c@{\hspace{3mm}}c@{\hspace{1mm}}@{\hspace{0.2mm}}}
			\toprule[1pt]
			\multirow{2}*{Methods}
			& \multicolumn{5}{@{\hspace{1mm}}c@{\hspace{1mm}}@{\hspace{0.2mm}}}{Perturbation with different noise levels} \\ \cline{2-6}
			& 0\% & 0.1\% & 0.5\% & 1.0\% & 2.0\%  \\ \hline \hline
			
			PU-Net~\cite{yu2018pu} &  0.699  & 0.717  & 0.794  & 0.860   & 0.945   \\
			MPU~\cite{yifan2018patch} & 0.348  & 0.364  & 0.426  & 0.524   & 0.831  \\
			PU-GAN~\cite{li2019pu} & 0.269  & 0.309  & 0.381  & 0.562   & 0.899    \\ \hline
			Our & \textbf{0.199} & \textbf{0.213}  & \textbf{0.229}  & \textbf{0.310}  & \textbf{0.592}    \\ \bottomrule[1pt]
	\end{tabular}}
\end{table}

\begin{table}[t]
	\caption{Comparing the upsampling performance of our full pipeline with various cases in the ablation study ($r$=16). Here we show CD values with the unit of  $10^{-3}$.}
	\centering
	\label{tab:ablation}
    \vspace{2mm}
	\resizebox{0.95\linewidth}{!}{%
		\begin{tabular}{c|cccc|c}
			\toprule[1pt]
			Model & Spatial Refiner  & Local & Global & Offset & CD\\
			\hline
			A &  &  &  & &  0.684  \\
			B & $\checkmark$  &    & $\checkmark$ & $\checkmark$& 0.378 \\
			C & $\checkmark$  &  $\checkmark$  &   & $\checkmark$& 0.343  \\
			D & $\checkmark$  & $\checkmark$   & $ \checkmark$ & & 0.228 \\
			\hline
			Full &     &    &   &    &\textbf{0.199} \\
			\bottomrule[1pt]
	\end{tabular}}
	\vspace*{-2mm}
\end{table}


\section{Conclusion}
\label{sec:conclusion}

In this paper, we present a disentangled refinement framework for point cloud upsampling.
Different from existing approaches that try to meet the various upsampling goals all in a single network, we propose to disentangle the upsampling task into two sub-goals, where we first generate coarse but dense points, and then refine these points by adjusting the location of each point.
To this end, we formulate an end-to-end disentangled refinement framework with two cascaded sub-networks: a dense generator and a spatial refiner.
In the spatial refiner, we introduce a pair of local and global refinement units to evolve the coarse feature map by considering both local and global geometric structures.
Also, we design our spatial refiner to regress offset vectors to adjust the coarse outputs in fine scale.
Experimental results demonstrate the superiority of our method over others.

Actually, this work aims to provide a generic framework to disentangle the point cloud upsampling task.
In the future, we may continue to explore a more comprehensive architecture for dense generator and spatial refiner.
We may further explore the possibility of designing a region adaptive refiner, meaning that we only fine-tune regions that are non-uniform and noisy, thus improving the overall efficiency.
Lastly, designing the refiner to be aware of edges may be helpful for downstream tasks like mesh reconstruction.
%

\para{Acknowledgments.} \
\new{We thank anonymous reviewers for the valuable comments.
This work is supported by the Hong Kong Centre for Logistics Robotics, and Research Grants Council of the Hong Kong Special Administrative Region (Project No. CUHK 14201717 \& 14201918 \&14201620).}

\newpage
\clearpage



{\small
\bibliographystyle{ieee_fullname}
\bibliography{egbib}

\begin{thebibliography}{10}\itemsep=-1pt

\bibitem{alexa2003computing}
Marc Alexa, Johannes Behr, Daniel Cohen-Or, Shachar Fleishman, David Levin, and
  Claudio~T. Silva.
\newblock Computing and rendering point set surfaces.
\newblock {\em IEEE Trans. Vis. $\&$ Comp. Graphics (TVCG)}, 9(1):3--15, 2003.

\bibitem{bernardini1999ball}
Fausto Bernardini, Joshua Mittleman, Holly Rushmeier, Cl{\'a}udio Silva, and
  Gabriel Taubin.
\newblock The ball-pivoting algorithm for surface reconstruction.
\newblock 5(4):349--359, 1999.

\bibitem{chen2020unpaired}
Xuelin Chen, Baoquan Chen, and Niloy~J. Mitra.
\newblock Unpaired point cloud completion on real scans using adversarial
  training.
\newblock {\em Int. Conf. on Learning Representations (ICLR)}, 2020.

\bibitem{chen2017multi}
Xiaozhi Chen, Huimin Ma, Ji Wan, Bo Li, and Tian Xia.
\newblock Multi-view {3D} object detection network for autonomous driving.
\newblock In {\em IEEE Conf. on Computer Vision and Pattern Recognition
  (CVPR)}, pages 1907--1915, 2017.

\bibitem{chen2020lassonet}
Zhutian Chen, Wei Zeng, Zhiguang Yang, Lingyun Yu, Chi-Wing Fu, and Huamin Qu.
\newblock {LassoNet}: Deep {L}asso-selection of {3D} point clouds.
\newblock {\em IEEE Trans. Vis. $\&$ Comp. Graphics (TVCG)}, 26(1):195--204,
  2020.

\bibitem{cole2006using}
David~M. Cole and Paul~M. Newman.
\newblock Using {Laser} range data for {3D} {SLAM} in outdoor environments.
\newblock In {\em IEEE Int. Conf. on Robotics and Automation (ICRA)}, pages
  1556--1563, 2006.

\bibitem{fan2017point}
Haoqiang Fan, Hao Su, and Leonidas~J. Guibas.
\newblock A point set generation network for {3D} object reconstruction from a
  single image.
\newblock In {\em IEEE Conf. on Computer Vision and Pattern Recognition
  (CVPR)}, pages 605--613, 2017.

\bibitem{he2016deep}
Kaiming He, Xiangyu Zhang, Shaoqing Ren, and Jian Sun.
\newblock Deep residual learning for image recognition.
\newblock In {\em Proceedings of the IEEE conference on computer vision and
  pattern recognition}, pages 770--778, 2016.

\bibitem{hermosilla2019total}
Pedro Hermosilla, Tobias Ritschel, and Timo Ropinski.
\newblock {Total Denoising}: Unsupervised learning of {3D} point cloud
  cleaning.
\newblock In {\em IEEE Int. Conf. on Computer Vision (ICCV)}, pages 52--60,
  2019.

\bibitem{huang2013edge}
Hui Huang, Shihao Wu, Minglun Gong, Daniel Cohen-Or, Uri Ascher, and Hao Zhang.
\newblock Edge-aware point set resampling.
\newblock {\em ACM Trans. on Graphics (TOG)}, 32(1):9:1--12, 2013.

\bibitem{lang2019pointpillars}
Alex~H. Lang, Sourabh Vora, Holger Caesar, Lubing Zhou, Jiong Yang, and Oscar
  Beijbom.
\newblock {PointPillars}: Fast encoders for object detection from point clouds.
\newblock In {\em IEEE Conf. on Computer Vision and Pattern Recognition
  (CVPR)}, pages 12697--12705, 2019.

\bibitem{li2019pu}
Ruihui Li, Xianzhi Li, Chi-Wing Fu, Daniel Cohen-Or, and Pheng-Ann Heng.
\newblock {PU-GAN}: A point cloud upsampling adversarial network.
\newblock In {\em IEEE Int. Conf. on Computer Vision (ICCV)}, pages 7203--7212,
  2019.

\bibitem{li2020pointaugment}
Ruihui Li, Xianzhi Li, Pheng-Ann Heng, and Chi-Wing Fu.
\newblock {PointAugment}: An auto-augmentation framework for point cloud
  classification.
\newblock In {\em IEEE Conf. on Computer Vision and Pattern Recognition
  (CVPR)}, pages 6378--6387, 2020.

\bibitem{li2020rotation}
Xianzhi Li, Ruihui Li, Guangyong Chen, Chi-Wing Fu, Daniel Cohen-Or, and
  Pheng-Ann Heng.
\newblock A rotation-invariant framework for deep point cloud analysis.
\newblock {\em arXiv preprint arXiv:2003.07238}, 2020.

\bibitem{li2020unsupervised}
Xianzhi Li, Lequan Yu, Chi-Wing Fu, Daniel Cohen-Or, and Pheng-Ann Heng.
\newblock Unsupervised detection of distinctive regions on 3d shapes.
\newblock {\em ACM Trans. on Graphics (TOG)}, 39(5):158:1--14, 2020.

\bibitem{li2018pointcnn}
Yangyan Li, Rui Bu, Mingchao Sun, Wei Wu, Xinhan Di, and Baoquan Chen.
\newblock {PointCNN}: Convolution on $\mathcal{X}$-transformed points.
\newblock In {\em Conference and Workshop on Neural Information Processing
  Systems (NeurIPS)}, pages 828--838, 2018.

\bibitem{lipman2007parameterization}
Yaron Lipman, Daniel Cohen-Or, David Levin, and Hillel Tal-Ezer.
\newblock Parameterization-free projection for geometry reconstruction.
\newblock {\em ACM Trans. on Graphics (SIGGRAPH)}, 26(3):22:1--5, 2007.

\bibitem{orts2016holoportation}
Sergio Orts-Escolano, Christoph Rhemann, Sean Fanello, Wayne Chang, Adarsh
  Kowdle, Yury Degtyarev, David Kim, Philip~L Davidson, Sameh Khamis, Mingsong
  Dou, et~al.
\newblock Holoportation: Virtual {3D} teleportation in real-time.
\newblock In {\em Proceedings of the 29th Annual Symposium on User Interface
  Software and Technology}, pages 741--754, 2016.

\bibitem{qi2019deep}
Charles~R. Qi, Or Litany, Kaiming He, and Leonidas~J. Guibas.
\newblock Deep hough voting for {3D} object detection in point clouds.
\newblock In {\em IEEE Int. Conf. on Computer Vision (ICCV)}, pages 9277--9286,
  2019.

\bibitem{qi2017pointnet}
Charles~R. Qi, Hao Su, Kaichun Mo, and Leonidas~J. Guibas.
\newblock {PointNet}: Deep learning on point sets for {3D} classification and
  segmentation.
\newblock In {\em IEEE Conf. on Computer Vision and Pattern Recognition
  (CVPR)}, pages 652--660, 2017.

\bibitem{qian2020pugeo}
Yue Qian, Junhui Hou, Sam Kwong, and Ying He.
\newblock {PUGeo-Net}: A geometry-centric network for {3D} point cloud
  upsampling.
\newblock In {\em European Conf. on Computer Vision (ECCV)}, 2020.

\bibitem{rakotosaona2019pointcleannet}
Marie-Julie Rakotosaona, Vittorio La~Barbera, Paul Guerrero, Niloy~J. Mitra,
  and Maks Ovsjanikov.
\newblock {PointCleanNet}: Learning to denoise and remove outliers from dense
  point clouds.
\newblock {\em Computer Graphics Forum (CGF)}, 39(1):185--203, 2020.

\bibitem{uy2019revisiting}
Mikaela~Angelina Uy, Quang-Hieu Pham, Binh-Son Hua, Thanh Nguyen, and Sai-Kit
  Yeung.
\newblock Revisiting point cloud classification: A new benchmark dataset and
  classification model on real-world data.
\newblock In {\em IEEE Int. Conf. on Computer Vision (ICCV)}, pages 1588--1597,
  2019.

\bibitem{wang2019deep}
Yue Wang and Justin~M. Solomon.
\newblock Deep closest point: Learning representations for point cloud
  registration.
\newblock In {\em IEEE Int. Conf. on Computer Vision (ICCV)}, pages 3523--3532,
  2019.

\bibitem{wu2015deep}
Shihao Wu, Hui Huang, Minglun Gong, Matthias Zwicker, and Daniel Cohen-Or.
\newblock Deep points consolidation.
\newblock {\em ACM Trans. on Graphics (SIGGRAPH Asia)}, 34(6):176:1--13, 2015.

\bibitem{yifan2018patch}
Wang Yifan, Shihao Wu, Hui Huang, Daniel Cohen-Or, and Olga Sorkine-Hornung.
\newblock Patch-based progressive 3{D} point set upsampling.
\newblock In {\em IEEE Conf. on Computer Vision and Pattern Recognition
  (CVPR)}, pages 5958--5967, 2019.

\bibitem{yu2018ec}
Lequan Yu, Xianzhi Li, Chi-Wing Fu, Daniel Cohen-Or, and Pheng-Ann Heng.
\newblock {EC-Net}: An edge-aware point set consolidation network.
\newblock In {\em European Conf. on Computer Vision (ECCV)}, pages 386--402,
  2018.

\bibitem{yu2018pu}
Lequan Yu, Xianzhi Li, Chi-Wing Fu, Daniel Cohen-Or, and Pheng-Ann Heng.
\newblock {PU-Net}: Point cloud upsampling network.
\newblock In {\em IEEE Conf. on Computer Vision and Pattern Recognition
  (CVPR)}, pages 2790--2799, 2018.

\bibitem{yuan2018pcn}
Wentao Yuan, Tejas Khot, David Held, Christoph Mertz, and Martial Hebert.
\newblock {PCN}: Point completion network.
\newblock In {\em Int. Conf. on {3D} Vision (3DV)}, pages 728--737, 2018.

\bibitem{zhang2019self}
Han Zhang, Ian Goodfellow, Dimitris Metaxas, and Augustus Odena.
\newblock Self-attention generative adversarial networks.
\newblock In {\em Int. Conf. on Machine Learning (ICML)}, pages 7354--7363.
  PMLR, 2019.

\bibitem{zhang2019shellnet}
Zhiyuan Zhang, Binh-Son Hua, and Sai-Kit Yeung.
\newblock Shell{N}et: Efficient point cloud convolutional neural networks using
  concentric shells statistics.
\newblock In {\em IEEE Int. Conf. on Computer Vision (ICCV)}, pages 1607--1616,
  2019.

\end{thebibliography}
}

\end{document}